\documentclass[11pt, letter]{article}
\usepackage[left=15mm, top= 1mm, bottom=1mm]{geometry}
\usepackage{amsmath}
\usepackage{amssymb}
\usepackage{amsfonts}
\usepackage{booktabs}
\usepackage{amsthm}
\usepackage{mathrsfs}
\usepackage[all]{xy}
\usepackage{multirow}
\usepackage{caption}
\usepackage{algorithm}
\usepackage{algpseudocode}
\usepackage{algorithm}
\usepackage{algcompatible}
\usepackage{graphicx}
\usepackage{subfigure}
\begin{document}
\title{A Modified Sigma-Pi-Sigma Neural Network with Adaptive Choice of Multinomials}


\author{Feng Li, Yan Liu, Khidir Shaib Mohamed, and Wei Wu
\thanks{W. Wu is the corresponding author (e-mail: wuweiw@dlut.edu.cn).
\newline \indent ~F. Li, K.S. Mohamed and W. Wu are with the School of Mathematical Sciences, Dalian University of Technology, Dalian 116024, China.
\newline \indent ~Y. Liu is with School of Information Science and Engineering, Dalian Polytechnic University, Dalian, China.}}

\date{}
\maketitle

\section* {ABSTRACT}
Sigma-Pi-Sigma neural networks (SPSNNs) as a kind of high-order neural networks can provide more powerful mapping capability than the traditional feedforward neural networks (Sigma-Sigma neural networks). In the existing literature, in order to reduce the number of the Pi nodes in the Pi layer, a special multinomial $P_s$ is used in SPSNNs. Each monomial in $P_s$ is linear with respect to each particular variable $ \sigma_i$ when the other variables are taken as constants. Therefore, the monomials like $\sigma_i^n$ or $\sigma_{i}^{n} \sigma_j$ with $n>1$ are not included. This choice may be somehow intuitive, but is not necessarily the best. We propose in this paper a modified Sigma-Pi-Sigma neural network (MSPSNN) with an adaptive approach to find a better multinomial for a given problem. To elaborate, we start from a complete multinomial with a given order. Then we employ a regularization technique in the learning process for the given problem to reduce the number of monomials used in the multinomial, and end up with a new SPSNN involving the same number of monomials (= the number of nodes in the Pi-layer) as in $P_s$. Numerical experiments on some benchmark problems show that our MSPSNN behaves better than the traditional SPSNN with $P_s$.

\section{Introduction }
Sigma-Pi-Sigma neural networks (SPSNNs) [1,4,7,8] as a kind of high-order neural networks can provide more powerful mapping capability [2-3,5,6] than the traditional feedforward neural networks (Sigma-Sigma neural networks). In an SPSNN, a Pi layer (denoted by $\Pi$ layer hereafter) is inserted in between the two Sigma layers. Each Pi node ($\Pi$ node) in the $\Pi$ layer corresponds to a monomial, of which the variables are the outputs of the Sigma nodes ($\Sigma$ nodes) of the first Sigma layer ($\Sigma_1$ layer). Each node in the second Sigma layer ($\Sigma_2$ layer) implements a linear combination of the outputs of the $\Pi$ layer, and therefore represents a multinomial expansion of the output $\sigma=(\sigma_1,\cdots,\sigma_N)$ of the $\Sigma_1$ layer. Then, the multinomial expansion is processed by an activation function in the $\Sigma_2$ layer to give the final output of the network.

At the beginning of the development of SPSNN, researchers have realized that it is not a good idea to include all the possible monomials in the $\Pi$ layer, i.e., to get a complete multinomial expansion of the $\Sigma_1$ layer, since it results in too many $\Pi$ nodes in the $\Pi$ layer. In the existing literature, in order to reduce the number of $\Pi$ nodes, a special multinomial $P_s$ (called multi-linear multinomial) is used in SPSNNs. The monomials in $P_s$ are linear with respect to each particular variable $\sigma_i$ when taking the other variables as constants. Therefore, the monomials such as $\sigma_{i}^n$ or $\sigma_i^n\sigma_j$ with $n>1$ are not included in $P_s$. An intuitive idea behind this strategy may be the following: A $\Pi$ node should receive at most one signal, rather than two or more signals, from each $\Sigma_1$ node.

But from general numerical approximation point of view, each monomial plays equally important role for approximating nonlinear mappings by using multinomial.  Thus, the special multi-linear multinomial $P_s$ may not be the best choice for the SPSNN to approximate a particular nonlinear mapping. To this end, we propose  an adaptive approach to find a better multinomial for a given problem. To elaborate, we start from a complete multinomial with a given order. Then we employ a regularization technique in the learning process for the given problem to reduce the number of monomials used in the multinomial, and end up with a modified SPSNN (MSPSNN) involving the same number of monomials (= the number of nodes in the $\Pi$ layer) as in $P_s$. In particular, a smoothing $L_{1/2}$ regularization term [10,15] is used as an example in our method, which has been successfully applied for various kinds of neural network regularization.

We divide the learning process of MSPSNN into two phases. The first phase is a structural optimization phase. Here, we  insert a regularization term   into the usual error function for SPSNN involving a complete set of multinomials, and perform a usual gradient learning process. In the end, we delete the $\Pi$ nodes with smaller $\Pi$-$\Sigma_2$ weights, and obtain a network with the same number of $\Pi$ nodes as in $P_s$.

The second learning phase is a refinement phase. We re-start a gradient learning process for the network obtained from the first learning phase, and use the  weights that survived the first phase as the initial weights. The aim of the refinement phase is to make up for the loss caused by the deleted nodes in the first learning phase.

Numerical experiments are performed on some benchmark problems including two approximation problems and two classification problems. It is shown that our new MSPSNN behaves  better than the traditional SPSNN with $P_s$.

The rest of the paper is arranged as follows. The proposed MSPSNN with smoothing $L_{1/2}$ regularization term is described in Section 2. In Section 3, Supporting numerical simulations are presented. Some conclusions are given in Section 4.

\section{ MSPSNN method with smoothing $L_{1/2}$ regularization}

\subsection{Sigma-Pi-Sigma neural network}

An SPSNN is composed of an input layer, two hidden layers of summation node layer ($\Sigma_1$ layer) and product node layer ($\Pi$ layer), and an output layer ($\Sigma_2$ layer). The numbers of nodes of these layers are $M+1, N, Q$ and 1, respectively.

Denote by $\mathbf{x}=(x_{0},\cdot\cdot\cdot,x_M )^T \in R^{M+1}$ the input vector, where the M components $x_{0},\cdots,x_{M-1}$ are the ``real" input, while $x_M$ is an extra artificial input, fixed to -1. The output vector $\sigma \in R^N$ of $\Sigma_1$ layer with respect to $\mathbf{x}$ can be written as
\begin{equation}\label{sigma}
\sigma=(\sigma_1,\cdots,\sigma_N)=(g(\mathbf{w}_{1}\cdot \mathbf{x}),g(\mathbf{w}_{2}\cdot \mathbf{x}),\cdot\cdot\cdot,g(\mathbf{w}_{N}\cdot \mathbf{x}))^T,
\end{equation}
where $g(\cdot)$ is a given nonlinear activation function, $\mathbf{w}_n=(w_{n0},\cdot\cdot\cdot,w_{nM} )^T \in R^{M+1} $ $(1\leq n\leq N)$ is the weight vector connecting the $n$-th summation node of $\Sigma_1$ layer and the input layer, and $\mathbf{w}_n\cdot \mathbf{x}$ denotes the inner product of $\mathbf{w}_n$ and $\mathbf{x}$. Here we remark that the component $w_{nM}$ usually represents the bias of the $n$-th summation node of $\Sigma_1$ layer.

In $\Pi$ layer, Each $\Pi$ node connects with certain nodes of $\Sigma_1$ layer, receives signals from these nodes, and outputs a particular monomial such as
\begin{equation}\label{example}
\sigma_1,\sigma_1\sigma_2,\sigma_1^2.
\end{equation}
Denote by $\land_q$ $ (1 \leq q \leq Q)$ the index set of all the nodes in $\Sigma_1$ layer that are connected to the $q$-th $\Pi$ node. For instance, let us assume that the above three examples in (\ref{example}) correspond to the first, third and fifth nodes of $\Pi$ layer, respectively. Then, we have
\begin{equation}
\land_1=\{1\}, \land_3=\{1,2\}, \land_5=\{1,1\}.
\end{equation}
The output vector $\tau=(\tau_1,\cdots,\tau_Q)^T\in R^{Q}$ of $\Pi$ layer is computed by
\begin{equation}\label{tao}
\mathbf{\tau}_q=\prod_{i\in \land_{q}}\sigma_{i}, 1\leq q\leq Q.
\end{equation}
Here we make a convention that $\tau_q=\prod_{i\in \land_{q}}\sigma_{i}\equiv 1$, when $\land_q=\phi$, i.e., when the $q$-th $\Pi$ node is not connected to any node of $\Sigma_1$ layer. The choice of $\land_{q}$'s is our main concern in this paper.

Before we concentrate our attention on the choice of $\land_{q}$'s, let us describe the output of $\Sigma_2$ layer.
 The output of the single node of $\Sigma_2$ layer, i.e., the final output of the network, is
\begin{equation}\label{y}
y=f(\mathbf{w}_{0}\cdot\tau),
\end{equation}
where $f(\cdot)$ is another given activation function, and $\mathbf{w}_{0}=(w_{0,1},w_{0,2},\cdots,w_{0,Q} )^T \in R^Q$ is the weight vector connecting $\Pi$ layer and $\Sigma_2$ layer. When the network is used for approximation problems, we usually set $f(t)=t$.
On the other hand, when the network is used for classification problems, $f(t)$ is usually chosen to be a Sigmoid function. In both the cases, we can see from (\ref{sigma}), (\ref{tao}) and (\ref{y}) that the input $\mathbf{w}_{0}\cdot\tau$ to $\Sigma_2$ layer is actually a multinomial expansion of the output values of $\Sigma_1$ layer, where the components of $\tau$ correspond to the monomials, and the components of $\mathbf{w}_{0}$ are the coefficients, involved in the multinomial expansion. As comparison, we recall that for the usual feedforward neural networks, the input to the $\Sigma_2$ layer is a linear combination of the output values of $\Sigma_1$ layer.

Now we discuss the choice of $\land_{q}$'s in detail and explain the main idea of the paper. For convenience and clarity, we take the third order multinomial of three variables as an example in this introduction section. Therefore, we have $N=3$, i.e., $\Sigma_1$ layer has three nodes.

We consider three choices of $\land_{q}$'s, resulting in three different multinomial expansions: the complete mutinomial, the partially linear multinomial (the traditional approach), and the adaptive multinomial (our proposed approach).

The choice of the complete mutinomial means that the input to $\Sigma_2$ layer is a complete multinomial as follows:
\begin{align}\label{threeordercomplete}
&w_{0,1}+w_{0,2}\sigma_{1}+w_{0,3}\sigma_{2}+w_{0,4}\sigma_{3}+w_{0,5}\sigma_{1}\sigma_{2}+w_{0,6}\sigma_{1}\sigma_{3}+w_{0,7}\sigma_{2}\sigma_{3}\nonumber\\
&+w_{0,8}\sigma_{1}^{2}+w_{0,9}\sigma_{2}^{2}+w_{0,10}\sigma_{3}^{2}+w_{0,11}\sigma_{2}\sigma_{1}^{2}+w_{0,12}\sigma_{3}\sigma_{1}^{2}+w_{0,13}\sigma_{1}\sigma_{2}^{2}\nonumber\\
&+w_{0,14}\sigma_{3}\sigma_{2}^{2}+w_{0,15}\sigma_{1}\sigma_{3}^{3}+w_{0,16}\sigma_{3}\sigma_{2}^{2}+w_{0,17}\sigma_{1}^{3}+w_{0,18}\sigma_{2}^{3}+w_{0,19}\sigma_{3}^{3}+w_{0,20}\sigma_{1}\sigma_{2}\sigma_{3}.
\end{align}
We see that there are twenty monomials in the multinomial expansion, corresponding to twenty $\Pi$ nodes in $\Pi$ layer. More generally, when $\Sigma_1$ layer has $N$ nodes, the number of the monomials is $C_{complete}^N=C_{N+3}^{3}$, which grows very rapidly when $N$ increases. Therefore, the complete multinomial approach is not a good choice in practice.

The traditional choice in the existing literature  is the partially linear multinomial approach: A partially linear multinomial is linear with respect to each particular variable $\sigma_i$, with the other variables taken as constants. For instance, the partially linear multinomial corresponds to (\ref{threeordercomplete}) is
\begin{equation}\label{eightterms}
w_{0,1}+w_{0,2}\sigma_{1}+w_{0,3}\sigma_{2}+w_{0,4}\sigma_{3}+w_{0,5}\sigma_{1}\sigma_{2}+w_{0,6}\sigma_{1}\sigma_{3}+w_{0,7}\sigma_{2}\sigma_{3}+w_{0,8} \sigma_{1}\sigma_{2}\sigma_{3}.
\end{equation}
We see that there are only eight monomials in (\ref{eightterms}), i.e., only eight nodes left in $\Pi$ layer. Generally, when $\Sigma_1$ layer has $N$ nodes, the number of the monomials is $C_{linear}^N=C_{N}^0+C_{N}^1+C_{N}^2+C_{N}^3$. Table \ref{tableNum} shows the comparison of $C_{complete}^N$ and $C_{linear}^N$ with different $N$. It can be seen that the difference becomes bigger when $N$ increases.

\begin{table}[hbpt]
\caption{Comparison of $C_{complete}^N$ and $C_{linear}^N$ with different $N$.}
\label{tableNum}
\centering
\begin{tabular}{ccccccccc}
\hline\noalign{\smallskip}
      $N$            &  3   &  4  &  5  &  6    &  7   &  8  &  9  &  10    \\
\noalign{\smallskip}\hline\noalign{\smallskip}
  $C_{complete}^N$   &  20  & 35  & 56  &  84   &  120 & 165 & 220 &  286  \\
\noalign{\smallskip}\hline
  $C_{linear}^N$     &  8   & 15  & 26  &  42   &  64  & 93  & 130 &  176  \\
\noalign{\smallskip}\hline
 Difference          &  12  & 20  & 30  &  42   &  56  & 72  & 90  &  110 \\
\noalign{\smallskip}\hline
\end{tabular}
\end{table}

The network structure corresponding to (\ref{eightterms}) is shown in Fig. \ref{fig:old8}. The corresponding $\land_q$'s are as follows:
\begin{equation}\label{landofeight}
\land_1=\{\phi\}, \land_2=\{1\},\land_3=\{2\},\land_4=\{3\},\land_5=\{1,2\}, \land_6=\{1,3\},\land_7=\{2,3\},\land_8=\{1,2,3\}.
\end{equation}
We observe that in (\ref{threeordercomplete}) and (\ref{eightterms}), the first product node, corresponding to the bias $w_{0,1}$, does not connect with any node in the $\Sigma_1$ layer, so $\land_1=\{\phi\}$. We also notice that there are no repeated indexes in each $\land_q$ in (\ref{landofeight}).

Our proposed choice is as follows: We start from a complete multinomial with a given order. Then we employ a regularization technique in the learning process to reduce the number of monomials used in the multinomial, and end up with a new SPSNN involving the same number of monomials as in the traditional choice. For instance, in the Example 1 given in Section 4.1 below, a new SPSNN is obtained with the following multimonial:
\begin{equation}
w_{0,1}+w_{0,2}\sigma_{1}+w_{0,3}\sigma_{2}\sigma_{3}+w_{0,4}\sigma_{1}\sigma_{2}^2+w_{0,5}\sigma_{2}\sigma_{3}^2+w_{0,6}\sigma_{1}^3+w_{0,7}\sigma_{2}^3+w_{0,8}\sigma_{3}^3.
\end{equation}
And correspondingly,
\begin{eqnarray}
\land_1=(\varnothing),\land_2=\{1\},\land_2=\{2,3\},\land_3=\{1,2,2\},\land_4=\{2,2,3\},\\
\land_5=\{2,3,3\},\land_6=\{1,1,1\},\land_7=\{2,2,2\},\land_8=\{3,3,3\}.\nonumber
\end{eqnarray}
We notice that now there are some repeated indexes in six $\land_q$'s.

\subsection{Error function with $L_{1/2}$ regularization}
Let the training samples be $\{\mathbf{x}^{j},O^{j}\}_{j=1}^{J}\subset R^{M+1} \times R $, where $\mathbf{x}^j =(x_{0}^j,\cdots,x_{M}^j )^T$ is the $j$-th input sample and $O^j$ is its corresponding ideal output. Let $y^j\in R$ $ (1\leq j \leq J)$ be the network output for the input $\mathbf{x}^j$. The aim of the training process is to build up a network such that the errors $ |y^j-O^j|$ $ (1 \leq j \leq J)$ are as small as possible.
A conventional square error function with no regularization term is as follows:
\begin{equation}
\tilde{E}(\mathbf{W})=\frac{1}{2}\sum_{j=1}^J(y^{j}-O^{j})^2=\sum_{j=1}^J g_{j}(\mathbf{w}_{0}\cdot\tau^{j}),
\end{equation}
where $\mathbf{W}=(\mathbf{w}_{0}^T,\mathbf{w}_{1}^T,\cdots,\mathbf{w}_{N}^T)$,
\begin{equation}
g_{j}(t)=\frac{1}{2}(g(t)-O^{j})^2,  t\in R, 1\leq j \leq J.
\end{equation}

Let us derive the gradient of the error function $\tilde{E}(\mathbf{W})$. Notice
\begin{equation}
\tau^{j}=(\tau_1^j,\tau_2^j,\cdots,\tau_Q^j)^{T}=(\prod_{i\in \land_1} \sigma_i^j,\prod_{i\in \land_2} \sigma_i^j,\cdot\cdot\cdot,\prod_{i\in \land_Q} \sigma_i^j)^{T}
\end{equation}
and
\begin{equation}
\mathbf{\sigma}^{j}=(\sigma_1^j,\sigma_2^j,\cdots,\sigma_N^j)^T=(g(\mathbf{w}_{1}\cdot \mathbf{x}^{j}),g(\mathbf{w}_{2}\cdot \mathbf{x}^{j}),\cdots,g(\mathbf{w}_{N}\cdot \mathbf{x}^{j}))^{T}.
\end{equation}
Then, the partial derivative of $\tilde{E}(\mathbf{W})$ with respect to $w_{0,q}$ $(1\leq q\leq Q)$ is
\begin{equation}
\tilde{E}_{w_{0,q}}(\mathbf{W})=\sum_{j=1}^J g'_j(\mathbf{w}_{0}\cdot \tau^{j})\tau_q^j.
\end{equation}
Moreover, for $1\leq n\leq N$, $0\leq m\leq M$ and $1\leq q\leq Q$, we have
\begin{equation}
\frac{\partial \tau_{q}}{\partial w_{nm}}=
\begin{cases}
(\prod_{i\in \land_q\backslash {n}} \sigma_{i})g'(\mathbf{w}_{n}\cdot \mathbf{x})x_{m}, & {\rm if}~ q\neq 1, ~{\rm and} ~n\in \land_q, \\
0, & {\rm if}~ q=1,~{\rm or} ~n\notin \land_q.
\end{cases}
\end{equation}
According to (4) and (16), for any $1\leq n\leq N,~0\leq m \leq M$, we have
\begin{align}
\tilde{E}_{w_{nm}}(\mathbf{W})&=\sum_{j=1}^J g'_j(\mathbf{w}_{0}\cdot \tau^{j})\sum_{q=1}^Q w_{0,q}\frac{\partial \tau_q^j}{\partial w_{nm}}\nonumber \\
&=\sum_{j=1}^J g'_j(\mathbf{w}_{0}\cdot \tau^{j})\sum_{q\in \bigvee_n} w_{0,q}(\prod_{i\in \land_Q\backslash {n}} \sigma_{i}^j)g'(\mathbf{w}_{n}\cdot \mathbf{x}^j)x_{m}^j,
\end{align}
where $\frac{\partial \tau_q^j}{\partial w_{nm}}$ denotes the value of $\frac{\partial \tau_q}{\partial w_{nm}}$ with $\sigma_{i}=\sigma_i^j$ and $\mathbf{x}=\mathbf{x}^{j}$ in (16).

The error function with the $L_{1/2}$ regularization term is
\begin{equation}\label{newerrorfunction}
E(\mathbf{W})=\tilde{E}(\mathbf{W})+\lambda[\sum_{q=1}^Q|w_{0,q}|^{1/2}+\sum_{n=1}^N(\sum_{m=0}^{M}|w_{nm}|)^{1/2}].
\end{equation}
The gradient method with $L_{1/2}$ regularization for training the network is: Starting with an arbitrary initial value $\mathbf{W}^0$, the weights $\{\mathbf{W}^k\}$ are updated iteratively by:
\begin{equation}
\mathbf{W}^{k+1}=\mathbf{W}^{k}-\triangle \mathbf{W}^{k}.
\end{equation}
Here, $\triangle \mathbf{W}^{k}=(\triangle w_{0,1}^{k},\cdots,\triangle w_{0,Q}^{k},\cdots,\triangle w_{10}^{k},\cdots,\triangle w_{NM}^{k})^T$ with
\begin{equation}
\triangle w_{0,q}^k=-\eta E_{w_{0,q}}(\mathbf{W}^k)=-\eta[\sum_{j=1}^J g'_j(\mathbf{w}_{0}^k\cdot \tau^{j})\tau_q^j+\frac{\lambda sgn(w_{0,q}^k)}{2|w_{0,q}^k|^{1/2}}]
\end{equation}
and
\begin{align}
\triangle w_{nm}^k&=-\eta E_{{w}_{nm}}(\mathbf{W}^{k})\nonumber \\
&=-\eta [\sum_{j=1}^J g'_j(\mathbf{w}_{0}^k\cdot \tau^{k,j})\sum_{q\in \lor_n} w_{0,q}^k(\prod_{i\in \land_Q\backslash {n}} \sigma_{i}^{k,j})g'(\mathbf{w}_{n}^{k}\cdot \mathbf{x}^j)x_{m}^j+\frac{\lambda sgn(w_{nm}^k)}{2(|w_{n0}^k|+\cdots+|w_{nm}^k|)^{1/2}}].
\end{align}
Here, $1\leq j\leq J; 1\leq n\leq N;0\leq m\leq M;1\leq q\leq Q;k=0,1,\cdots;$ $\eta> 0$ is the learning rate; and $ \lambda>0$ is the regularization parameter.

\subsection{Error function with smoothing $L_{1/2}$ regularization}
We note that the usual $L_{1/2}$ regularization term in (18) is a non-differentiable function at the origin. In previous studies [][], it has been replaced by a smoothing function as follows
\begin{equation}
E(\mathbf{W})=\tilde{E}(\mathbf{W})+\lambda[\sum_{q=1}^Q|f(w_{0,q})|^{1/2}+\sum_{n=1}^N(\sum_{m=0}^M|f(w_{nm})|)^{1/2}],
\end{equation}
where $f(x)$ is the following piecewise multinomial function:
\begin{equation}
f(x)=
\begin{cases}
|x|, & if~ |x| \geq a,\\
-\frac{x^{4}}{8a^{3}}+\frac{3x^{2}}{4a}+\frac{3a}{8}, & if~ |x| < 0.
\end{cases}
\end{equation}
It is easy to obtain that
\begin{equation}
f(x)\in [\frac{3a}{8},+\infty),~f'(x)\in[-1,1],~{\rm and}~f''(z)\in[0,\frac{3}{2a}].\nonumber
\end{equation}
The gradient of the error function can be written as
\begin{equation}
E_{\mathbf{W}}(\mathbf{W})=(E_{w_{0,1}}(\mathbf{W}),E_{w_{0,2}}(\mathbf{W}),\cdots,E_{w_{0,Q}}(\mathbf{W}),E_{w_{10}}(\mathbf{W}),E_{w_{11}}(\mathbf{W}),\cdots,E_{w_{NM}}(\mathbf{W}))^T,
\end{equation}
where
\begin{align*}
&E_{w_{0,q}}(\mathbf{W})=\sum_{j=1}^J g'_j(\mathbf{w}_{0}\cdot \tau^{j})\tau_q^j+\frac{\lambda f'(w_{0,q})}{2(f(w_{0,q}))^{1/2}}\\
&E_{w_{nm}}(\mathbf{W})= \sum_{j=1}^J g'_j(\mathbf{w}_{0}\cdot \tau^{j})\sum_{q\in \lor_n} w_{0,q}(\prod_{i\in \land_Q\backslash {n}} \sigma_{i}^{j})g'(\mathbf{w}_{n}\cdot \mathbf{x}^j)x_{m}^j+\frac{\lambda f'(w_{0,q})}{2(f(w_{n0})+\cdots+f(w_{nm}))^{1/2}}).
\end{align*}
Starting from an arbitrary initial value $\mathbf{W}^0$, the gradient method with the  smoothing $L_{1/2}$ regularization updates the weights $\{\mathbf{W}^k \}$ iteratively by
\begin{equation}
\mathbf{W}^{k+1}=\mathbf{W}^{k}-\triangle \mathbf{W}^{k}
\end{equation}
with
\begin{align}
 \triangle w_{0,q}^k&=-\eta E_{w_{0,q}}(\mathbf{W}^{k})
 =-\eta [\sum_{j=1}^J g'_j(\mathbf{w}_{0}^{k}\cdot \tau^{k,j})\tau_q^j+\frac{\lambda f'(w_{0,q}^{k})}{2(f(w_{0,q}^{k}))^{1/2}}]
\end{align}
and
\begin{align}
 &\triangle w_{nm}^k =-\eta E_{nm}(\mathbf{W}^{k})\nonumber \\
 &=-\eta [\sum_{j=1}^J g'_j(\mathbf{w}_{0}^{k}\cdot \tau^{k,j})\sum_{q\in \lor_n} w_{0,q}^{k}(\prod_{i\in \land_Q\backslash {n}} \sigma_{i}^{k,j})g'(\mathbf{w}_{n}^{k}\cdot \mathbf{x}^j)x_{m}^j+\frac{\lambda f'(w_{nm}^{k})}{2(f(w_{n0}^{k})+\cdots+f(w_{nM}^{k}))^{1/2}}],
\end{align}
where $1\leq j\leq J;~1\leq n\leq N;~0\leq m\leq M;~1\leq q\leq Q;~k=0,1,\cdots;~\eta> 0$ is the learning rate; and $ \lambda>0$ the regularization parameter.

\section{Algorithm}
As mentioned in the Introduction, We divide the learning process into two phases: a structural optimization phase for choosing the structure of the network, followed by a refinement phase for finally  choosing the weights. Detailed descriptions of these two training phases are given in the following Algorithms 1 and 2, respectively.

\begin{algorithm}[H]
\caption{Structural optimization}
\begin{algorithmic}
\STATE {\bf Input.} {Input the dimension $M$, the number $N$ of the $\Sigma_1$ nodes, the number $Q$ of the $\Pi$ nodes, the maximum iteration number $I$, the learning rate $\eta$, the regularization parameter $\lambda$, and the training samples $\{\mathbf{x}^j,O^j \}_{j=1}^J \subset R^{M+1}\times R$.}
\STATE {\bf Initialization.} Initialize randomly the initial weight vectors $\mathbf{w}_0^0=(w_{0,1}^0,\cdots,w_{0,Q}^0 )^T \in R^Q$ and $\mathbf{w}_{n}^0=(w_{n0}^0,w_{n1}^0,\cdots,w_{nM}^0 )^{T}\in R^{M+1} ~(1\leq n \leq N)$.
\STATE {\bf Training.} {${\bf  For}~ k=1,2,\cdot\cdot\cdot,I ~ {\bf do}$ \\
~~~Compute the error function (22).\\
~~~Compute  the gradients (26) and (27).\\
~~~Update the weights $\mathbf{w}_0^k$ and $\mathbf{w}_n^k~(1\leq n \leq N)$ by using (25).\\
{\bf end}}
\STATE {\bf Structural optimization.} In the obtained weight vector  $\mathbf{w}_0^I=(w_{0,1}^I,\cdots,w_{0,Q} )^T$, select the $\hat{Q}=C_{linear}^N$ largest weights in absolute value to form a vector  $\mathbf{\hat{w}_0}=\{\hat{w}_1,\hat{w}_2,\cdots,\hat{w}_{\hat{Q}}\}$.
\STATE {\bf Output.} Output the final weight vectors $\mathbf{\hat{w}_0}$ and $\mathbf{\hat{w}_n}=\mathbf{{w}_n^I}$ $(1\leq n\leq N)$.
\algstore{testcont}
\end{algorithmic}
\end{algorithm}

\begin{algorithm}[H]
\caption{Refinement training}
\begin{algorithmic}
\algrestore{testcont}
\STATE  {\bf Input.} {Input the dimension $M$,  the  number $N$ of the  $\Sigma_1$ nodes, the number $\hat{Q}$ of the $\Pi$ nodes, the maximum iteration number $K$, the learning rate $\eta$,  and the training samples $\{\mathbf{x}^j,O^j \}_{j=1}^J \subset R^{M+1} \times R$. }
\STATE {\bf Initialization.} {Set $\mathbf{w}_0^0=\mathbf{\hat{w}_0}$ and $w_{n}^0=\mathbf{\hat{w}_n}$ $(1\leq n \leq N)$, and $\lambda=0$.}
\STATE {\bf Refinement Training.} ${\bf  for}~ k=1,2,\cdots,K ~ {\bf do}$ \\
~~~Compute the error function by (22).\\
~~~~Compute the gradient of the weights $\bigtriangleup w_{0}^k$ and $\bigtriangleup w_n^k~(1\leq n \leq N)$ by (26) and (27), respectively.\\
~~~Update the weights $w_0^k$ and $w_n^k$ $  (1\leq n \leq N)$ by (25).\\
{\bf end}\\
{\bf Output.} Output the final weight vectors $\mathbf{w}_0^K$ and $\mathbf{w}_n^K~(1\leq n\leq N)$.
\end{algorithmic}
\end{algorithm}

\section{Numerical experiments}
In this section, the proposed method is performed on four numerical benchmark problems: Mayas' function problem, Gabor function problem, Sonar problem and the Pima Indians diabetes data classification with different learning rates.

\subsection{Example 1: Mayas' function approximate}
In this example, a network is considered to approximate the Mayas' function as below:
\begin{equation}
H_{M}(x,y)=0.26(x^2+y^2)-0.48xy.
\end{equation}
The training samples of the network are 36 input points selected from an even $6\times 6$ grid on $-0.5\leq x\leq 0.5$ and $-0.5\leq y\leq 0.5$. Similarly, the test samples are 400 input points selected from $20\times20$ grid on $-0.5\leq y\leq 0.5$ and $-0.5\leq y \leq 0.5$.

\begin{figure}[!hbpt]
\centering
\vspace{-1em}
\subfigure[]{\label{fig:old8}\includegraphics[width=2.5in]{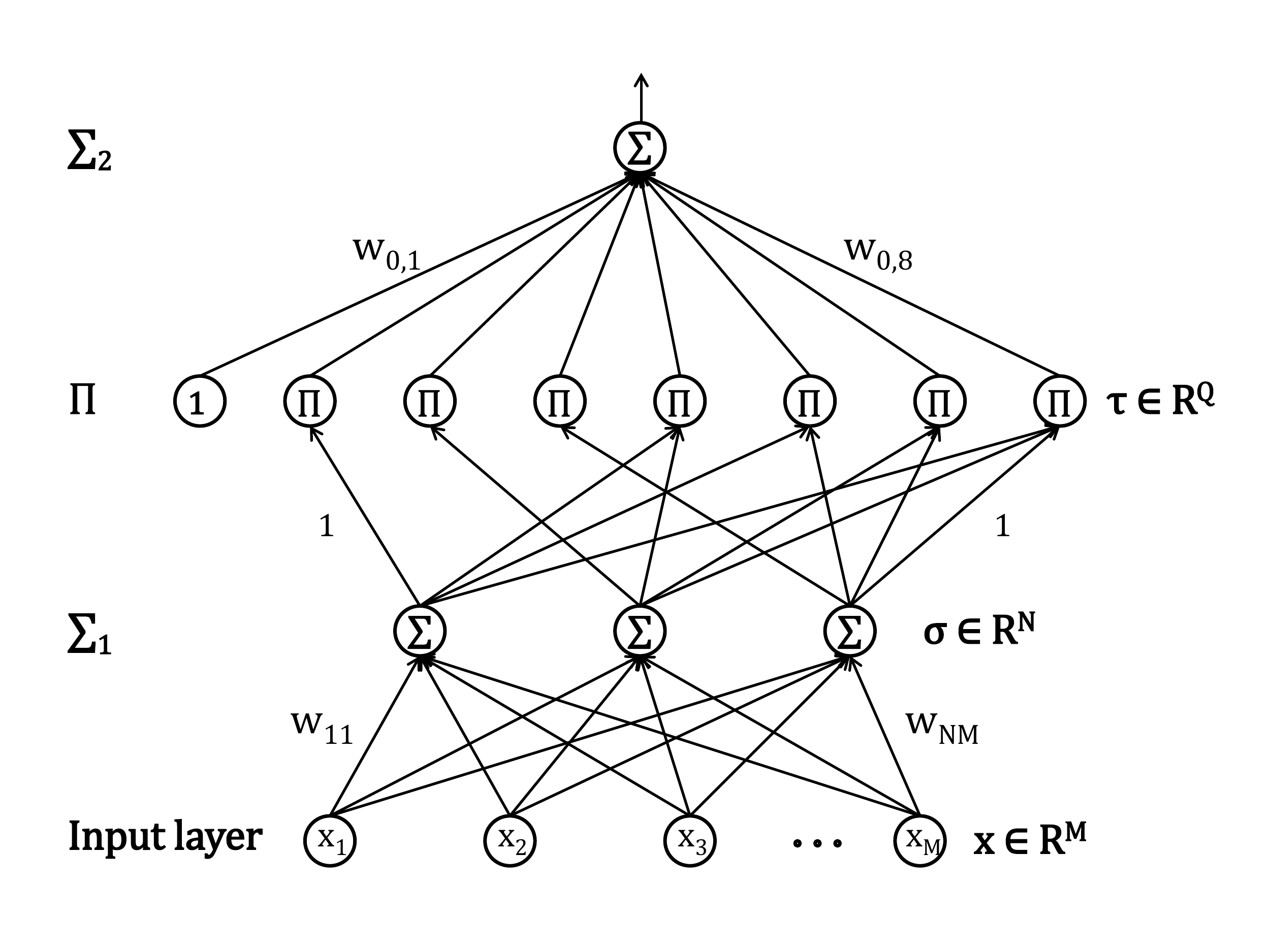}} \hspace{-0.3in}
\vspace{-1em}
\subfigure[]{\label{fig:new8Mayas}\includegraphics[width=2.5in]{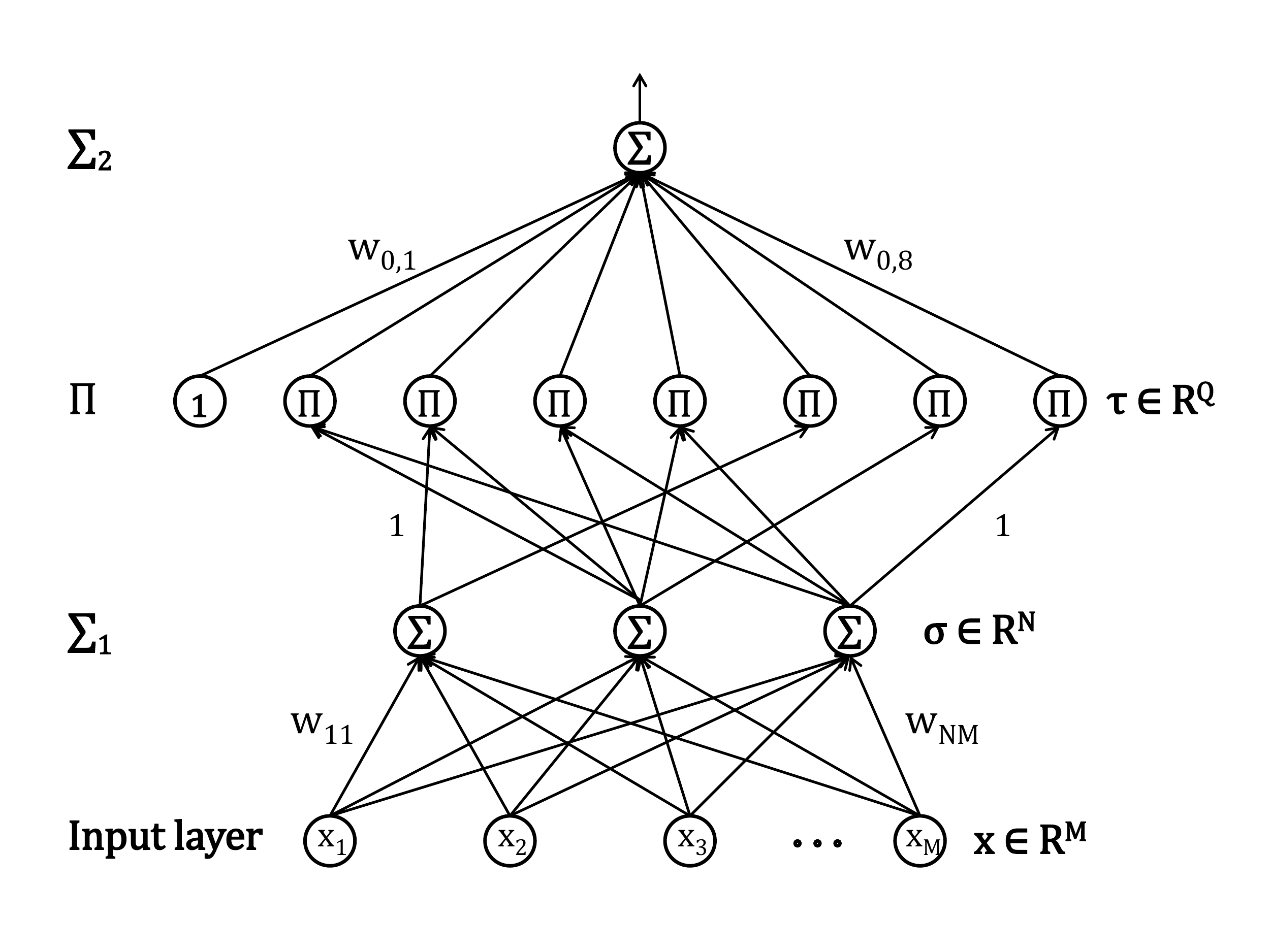}}
\vspace{-1em}
\subfigure[]{\label{fig:new8Gabor}\includegraphics[width=2.5in]{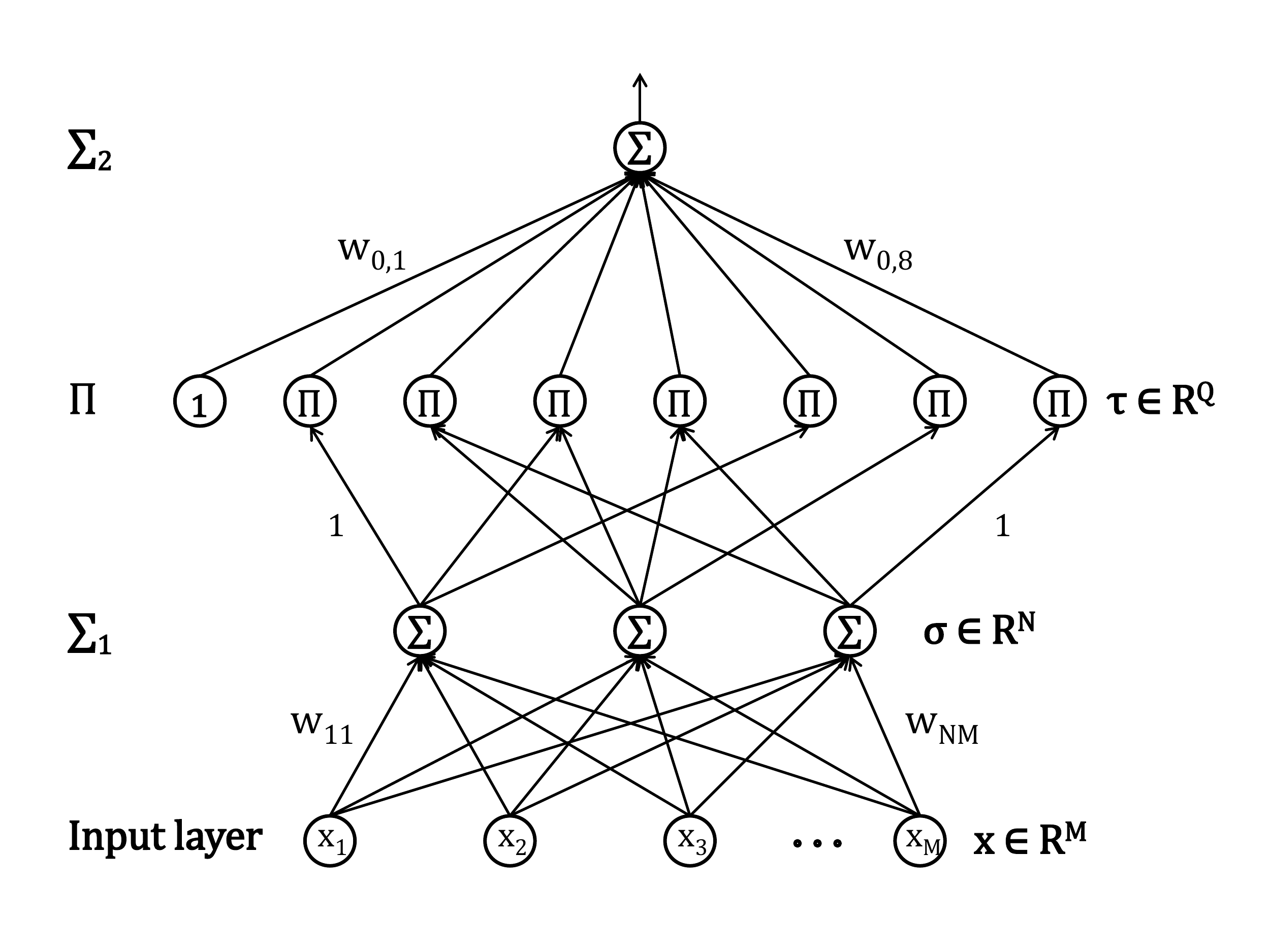}} \hspace{-0.3in}
\subfigure[]{\label{fig:new8Sonar}\includegraphics[width=2.5in]{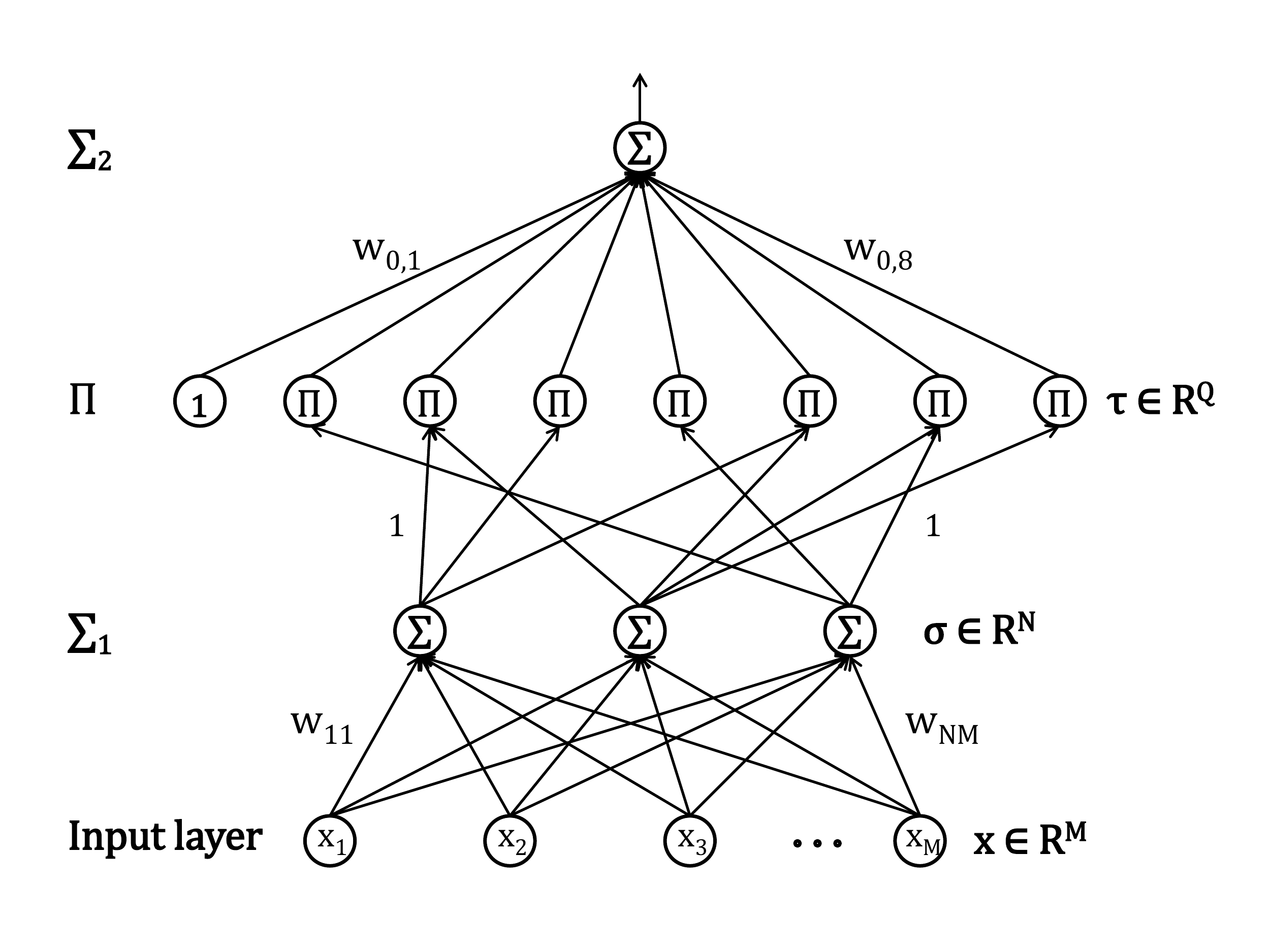}}
\caption{(a) Old structure; (b) New structure based on Mayas' function approximation; (c) New structure based on Gabor function approximation; (d) New structure based on Sonar problem approximation.} \label{fig:8}
\end{figure}

After performing Algorithms 1 with $\eta=0.005$, $\lambda=0.0001$ and $iteration_{max}=5000$, we select eight monomials, $1,~\sigma_{2}\sigma_{3},~\sigma_{1}\sigma_{2}^2,~\sigma_{3}\sigma_{2}^2,~\sigma_{1}^3,~\sigma_{2}^3,~\sigma_{3}^3$, to approximate the complete multinomial. The new structure corresponds to Fig. \ref{fig:new8Mayas}. The new weighted linear combination is
\begin{equation}
w_{0,1}+w_{0,2}\sigma_{1}+w_{0,3}\sigma_{2}\sigma_{3}+w_{0,4}\sigma_{1}\sigma_{2}^2+w_{0,5}\sigma_{2}\sigma_{3}^2+w_{0,6}\sigma_{1}^3+w_{0,7}\sigma_{2}^3+w_{0,8}\sigma_{3}^3
\end{equation}
From Fig. \ref{fig:new8Mayas}, the first product node, corresponding to the bias $w_{0,1}$, does not connect with any node in the $\Sigma_1$  layer, so $\land_1=\phi.$ And we have
\begin{align}
&\land_1=\varnothing,\land_2=\{1\},\land_2=\{2,3\},\land_3=\{1,2,2\},\land_4=\{2,2,3\}...\nonumber\\
&\land_5=\{2,3,3\},\land_6=\{1,1,1\},\land_7=\{2,2,2\},\land_8=\{3,3,3\}
\end{align}

\begin{figure}[ht]
\vspace{-1em}
\centering
\includegraphics[width=3.7in]{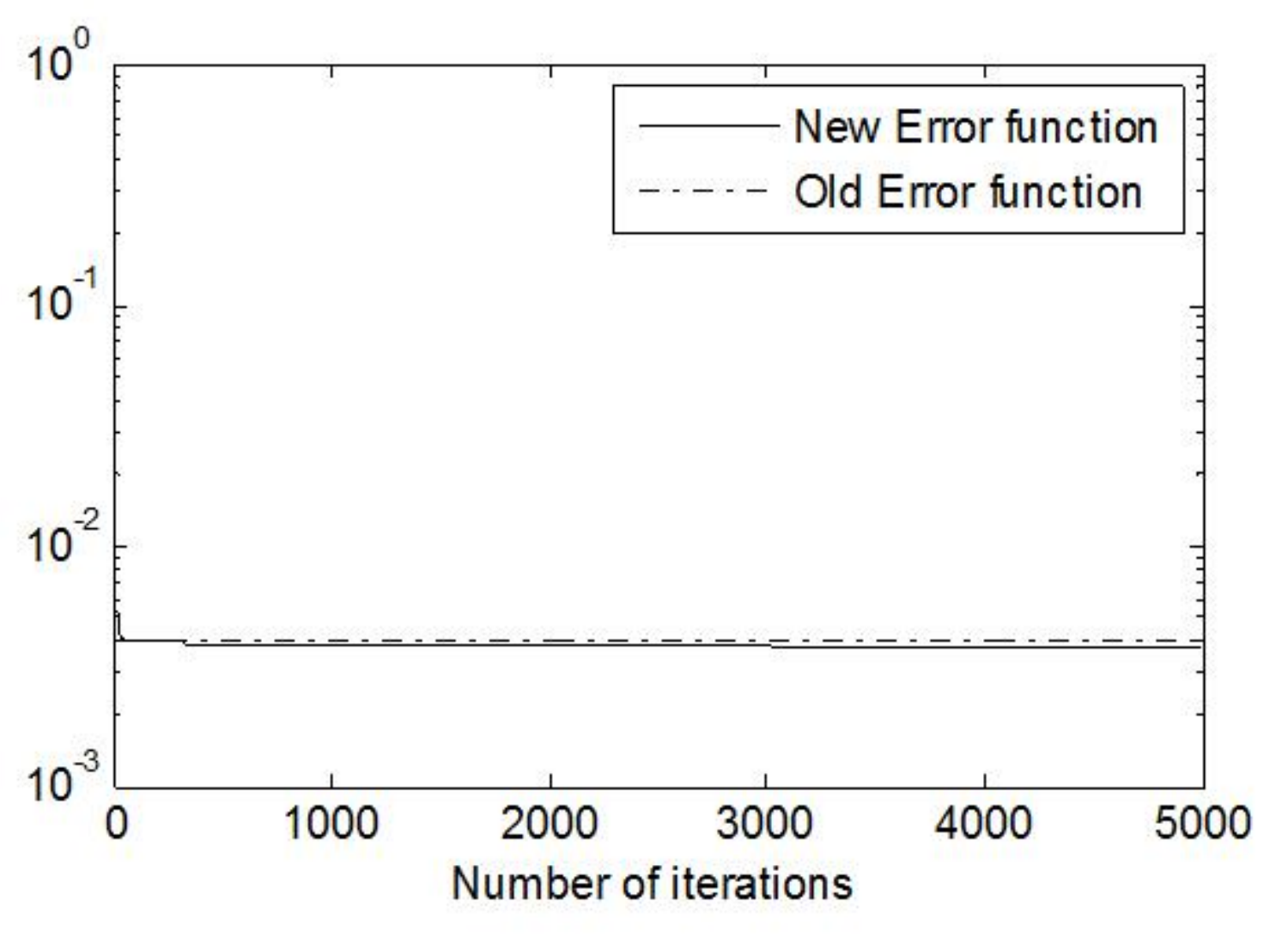}
\vspace{-1em}
\caption{Comparison of error for Mayas¡¯ approximation problem.}
\label{fig:Mayaserror}
\end{figure}

Then, we perform Algorithms 2 and use the test samples to evaluate our method. The average error with different parameter $\eta$ over the 20 tests and the improvement of the performance have been shown in Table \ref{tab:Mayas}. The persuasive comparison shows that the new structure attains the best effectiveness, i.e., the smallest error. From Fig. \ref{fig:Mayaserror}, we see that the surface of Mayas' error function in new structures is monotonically decreasing and converge to 0. 

\begin{table}[!ht]
  \centering
  \caption{Comparison of average error for Mayas' approximate error function.}
  \label{tab:Mayas}
  \begin{tabular}{cccc}
    \toprule
    Learning Rate& Average Old & Average New& Improvement \%\\
    \midrule
    0.001 & 0.0042 & 0.0041 & 2.38\\
    0.005 & 0.0043 & 0.0040 & 6.98 \\
    0.01  & 0.0040 & 0.0039 & 2.5 \\
    0.05  & 0.0039 & 0.0033 & 15.38\\
    0.1   & 0.0040 & 0.0035 & 12.5 \\
    \bottomrule
  \end{tabular}
\end{table}

\subsection{Example 2: Gabor function approximate}
In this example, a MPSPNN is used to approximate the Gabor function as:
\begin{equation}
H_{G}=\frac{1}{2\pi(0.5)^{2}}exp\left(\frac{x^{2}+y^{2}}{2(0.5)^{2}}\right)cos(2\pi(x+y))
\end{equation}
The training samples of the neural network are 36 input points selected from an evenly $6\times 6$ grid on $-0.5\leq x\leq 0.5$ and $-0.5\leq y\leq 0.5.$ Similarly, the test samples are 400 input points selected from $20\times 20$ grid on $-0.5\leq y\leq 0.5$ and $-0.5\leq y\leq 0.5$.
By performing Algorithms 1 with $\eta=0.009$, $\lambda=0.0001$ and $iteration_{max}=5000$, $1,\sigma_{1},\sigma_{2}\sigma_{3},\sigma_{1}\sigma_{2}^{2},\sigma_{2}\sigma_{3}^{2},\sigma_{1}^{3},\sigma_{2}^{3},$ and $\sigma_{3}^{3}$ are selected to approximate the complete multinomial. The new structure corresponds to Fig. \ref{fig:new8Gabor} and the new weighted linear combination is
\begin{equation}
w_{0,1}+w_{0,2}\sigma_{1}+w_{0,3}\sigma_{2}\sigma_{3}+w_{0,4}\sigma_{1}\sigma_{2}^{2}+w_{0,5}\sigma_{2}\sigma_{3}^{2}+w_{0,6}\sigma_{1}^{3}+w_{0,7}\sigma_{2}^{3}+w_{0,8}\sigma_{3}^{3}
\end{equation}
and we have
\begin{equation}
\land_1=\varnothing, \land_2=\{1\},\land_3=\{2,3\},\land_4=\{1,2,2\},\land_5=\{2,3,3\},\land_6=\{1,1,1\},\land_7=\{2,2,2\}, \land_8=\{3,3,3\}
\end{equation}

\begin{figure}[ht]
\vspace{-1em}
\centering
  \includegraphics[width=3.7in]{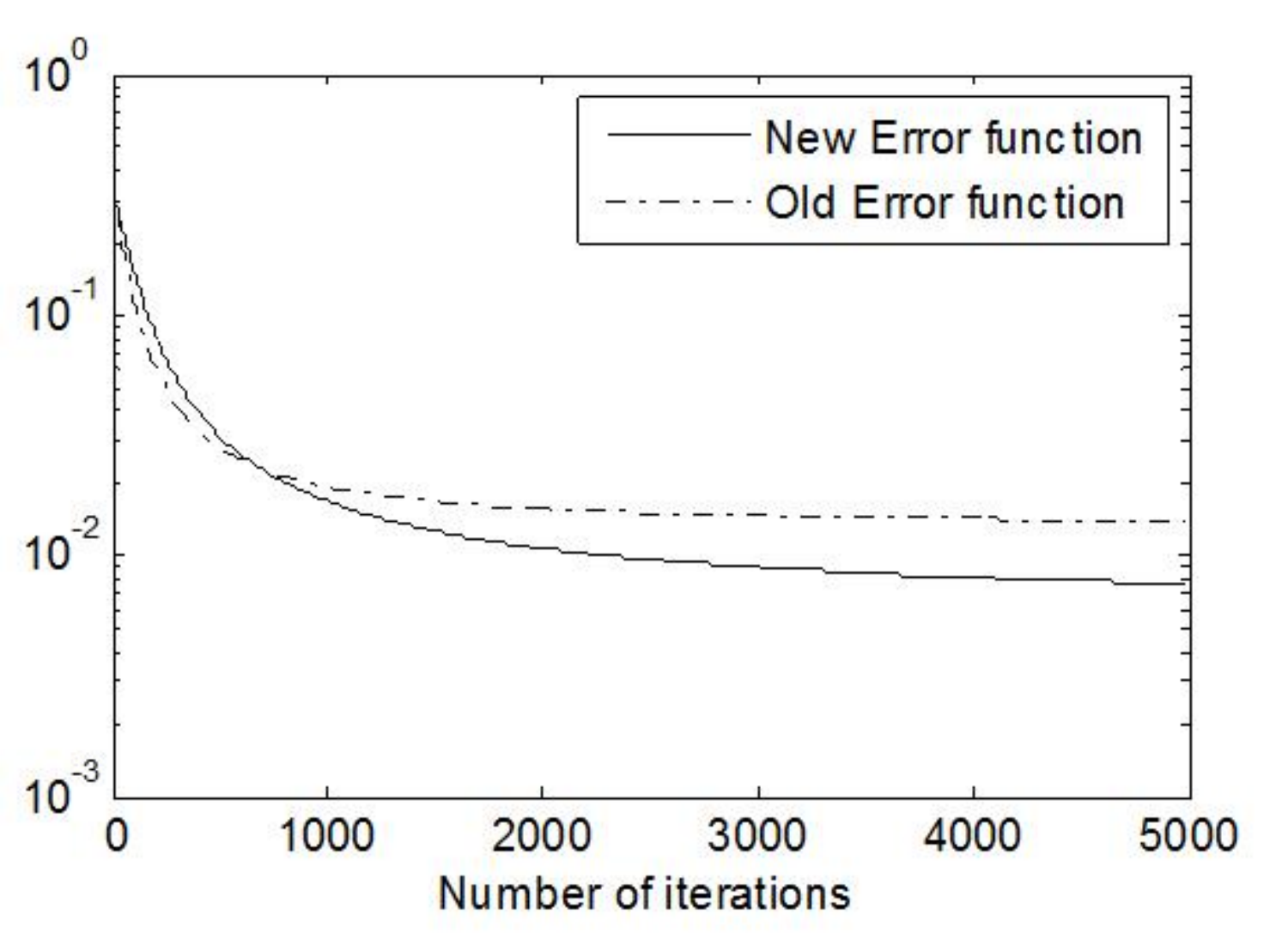}
\vspace{-1em}
\caption{Comparison of error for Gabor approximation problem.}
\label{fig:Gaborerror}
\end{figure}

Then, we perform Algorithms 2 and use the test samples to evaluate our method. The average error and the improvement of the performance have been shown in Table \ref{tab:Gabor}. The results show that the new structure attains the smallest error. From Fig. \ref{fig:Gaborerror}, we see that the surface of Gabor error function in new structures is monotonically decreasing and converge to 0, as predicted by Theorem 1.

\begin{table}[!ht]
  \centering
  \caption{Comparison of average error for Gabor approximate error function.}
  \label{tab:Gabor}
  \begin{tabular}{cccc}
    \toprule
    Learning Rate& Average Old & Average New& Improvement \%\\
    \midrule
    0.001 & 0.0131 & 0.0075 & 42.75\\
    0.005 & 0.0133 & 0.0065 & 51.13 \\
    0.01  & 0.0130 & 0.0064 & 50.77 \\
    0.05  & 0.0132 & 0.0063 & 52.27\\
    0.1   & 0.0131 & 0.0055 & 58.02 \\
    \bottomrule
  \end{tabular}
\end{table}

\subsection{Example 3: Sonar data classification}
Sonar problem is a well-known benchmark dataset, which aims to classify reflected sonar signals into two categories (metal cylinders and rocks). The related data set comprises 208 input vectors, each with 60 components. In this example, 4-fold cross validation is used to perform experiments, that is, $75\%$ samples for training and $25\%$ samples for testing are stochastically selected from the 208 samples. After performing our method, $1,\sigma_{3},\sigma_{1}\sigma_{2},\sigma_{1}^{2},\sigma_{3}^{2},\sigma_{2}\sigma_{1}^{2},\sigma_{2}\sigma_{3}^{2}$  and $\sigma_{2}^{3}$ are selected to to approximate the complete multinomial. The new structure corresponds to Fig. \ref{fig:new8Sonar} and the new weighted linear combination is
\begin{equation}
w_{0,1}+w_{0,2}\sigma_{3}+w_{0,3}\sigma_{1}\sigma_{2}+w_{0,4}\sigma_{1}^{2}+w_{0,5}\sigma_{3}^{2}+w_{0,6}\sigma_{2}\sigma_{1}^{2}+w_{0,7}\sigma_{2}\sigma_{3}^{2}+w_{0,8}\sigma_{2}^{3}
\end{equation}
Then, we have
\begin{equation}
\land_1=\varnothing, \land_2=\{2\},\land_3=\{1,2\},\land_4=\{1,1\},\land_5=\{3,3\},\land_6=\{1,1,2\},\land_7=\{2,3,3\}, \land_8=\{2,2,2\}
\end{equation}

\begin{table}[!ht]
\centering
\caption{Comparison of average classification accuracy for sonar problem.}
\label{tab:table3}
\begin{tabular}{ccccccc}
\toprule
   Round& Old Train& New Train & Impprovment\% &Old Test & New Test & Improvement\%\\
\midrule
    1  & 79.42 & 89.26 & 11.67 & 71.16 & 83.18 & 15.58 \\
    2  & 86.22 & 95.67 & 10.39 & 81.02 & 90.74 & 11.32 \\
    3  & 85.12 & 87.51 & 2.77  & 78.85 & 81.73 & 3.59 \\
    4  & 80.77 & 90.87 & 11.77 & 68.52 & 82.85 & 18.93 \\
    5  & 79.97 & 84.94 & 6.03  & 76.28 & 80.45 & 5.32 \\
\bottomrule
    Overall  & 82.30 & 89.65 & 8.67 & 75.17 & 83.79 &10.85 \\
\bottomrule
\end{tabular}
\end{table}
\begin{table}[!ht]
  \centering
  \caption{Comparison of the best classification accuracy for sonar problem.}
  \label{tab:table4}
  \begin{tabular}{ccccccc}
    \toprule
   Round& Old Train& New Train & Impprovment\% &Old Test & New Test & Improvement\%\\
    \midrule
    1  & 89.47 & 99.36 & 10.48  & 84.62 & 98.11 & 14.76 \\
    2  & 94.87 & 99.36 & 4.62   & 89.47 & 95.62 & 6.65 \\
    3  & 95.62 & 96.15 & 0.55   & 88.46 & 92.31 & 4.26 \\
    4  & 84.62 & 100.0 & 16.66  & 89.47  & 100.0 & 11.12 \\
    5  & 88.46 & 100.0 & 12.25  & 79.49 & 88.46 & 10.65 \\
    \bottomrule
    Overall  & 90.61& 98.97& 8.82 & 86.30 & 94.90 &9.49 \\
    \bottomrule
  \end{tabular}
\end{table}
\begin{table}[!ht]
  \centering
  \caption{Comparison of the worst classification accuracy for sonar problem.}
  \label{tab:table5}
  \begin{tabular}{ccccccc}
    \toprule
   Round& Old Train& New Train & Impprovment\% &Old Test & New Test & Improvement\%\\
    \midrule
    1  & 71.79 & 80.77 & 11.77  & 57.69 & 69.23 & 18.18 \\
    2  & 73.08 & 91.03 & 21.88  & 71.79 & 84.62 & 16.41 \\
    3  & 73.08 & 78.21 & 6.85   & 61.54 & 65.38 & 6.05 \\
    4  & 71.79 & 75.00 & 4.37   & 50.0  & 71.79 & 35.78 \\
    5  & 75.00 & 78.21 & 4.19   & 71.79 & 78.21 & 8.56 \\
    \bottomrule
    Overall  & 72.95 & 80.64 & 10.01 & 62.56 & 73.85 & 16.55 \\
    \bottomrule
  \end{tabular}
\end{table}

In both structures, 20 trials are carried out for each learning algorithm. In Tables 4, 5 and 6, we compare average accuracy, best accuracy and worst accuracy of classification of both structures, respectively. In all three tables, it can be seen that new structure is more advantageous than the old structure. These show that our new structure is better and monotonically decreasing and converge to 0 during the iterative learning as predicted by Theorem 1.
\subsection{Example 4: Pima Indians diabetes data classification}
To verify the theoretical evaluation of MSPSNNs, we used a Pima Indians Diabetes Database, which comprises 768 samples with 8 attributes. The dataset is available at UCI machine learning repository (http://archive.ics.uci.edu/ml/datasets/Pima+Indians+Diabetes). 4-fold cross validation is used to perform our method.

\begin{figure}[ht]
  \centering
  \includegraphics[width=4in]{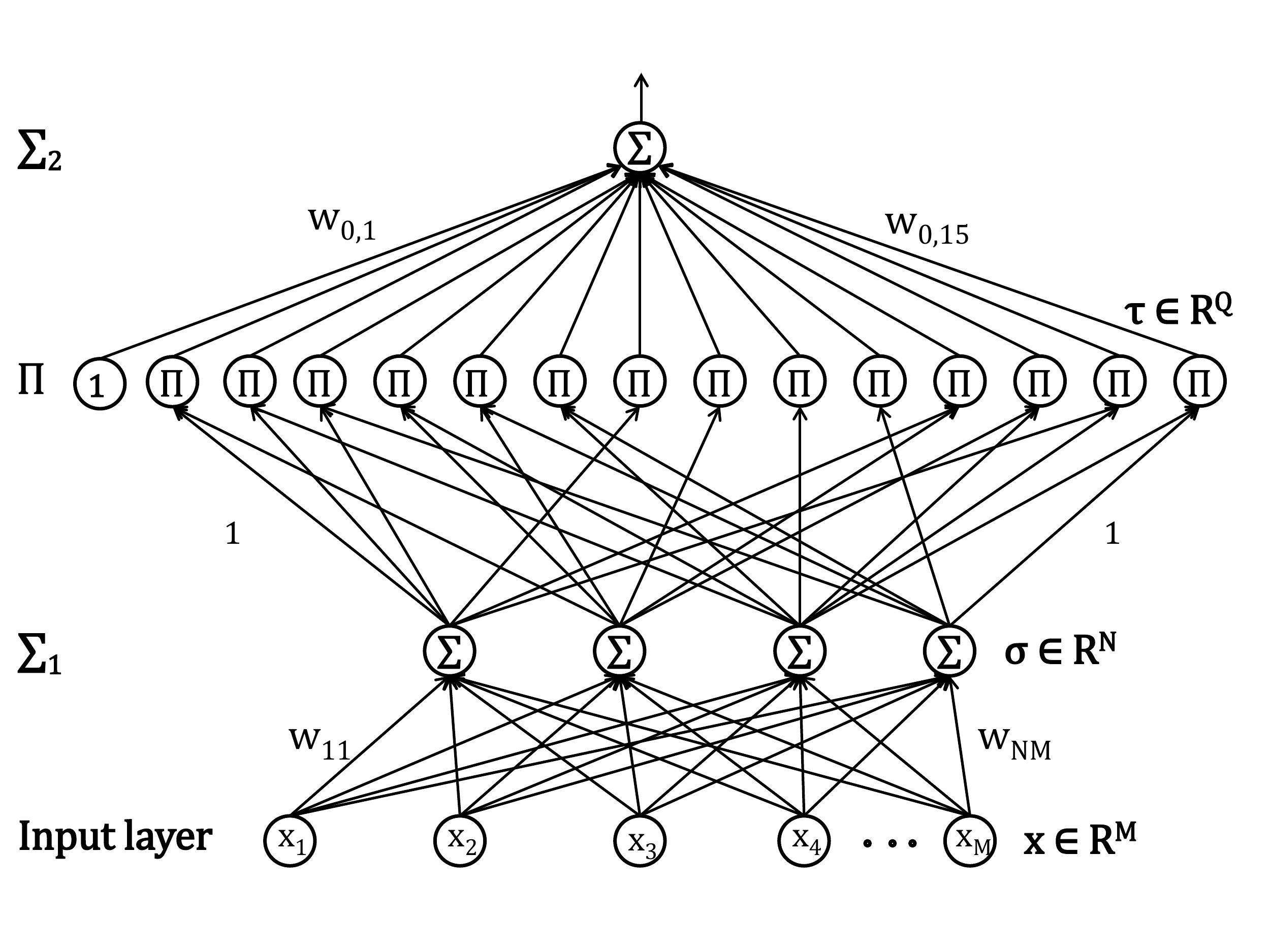}
  \caption{Four-input three-order new structure based on Pima Indians diabetes data classification.}
  \label{fig:new15}
\end{figure}

After that, $1,\sigma_{1}\sigma_{2},\sigma_{1}\sigma_{3},\sigma_{1}\sigma_{4},\sigma_{2}\sigma_{3},\sigma_{2}\sigma_{4},\sigma_{3}\sigma_{4}$
$,\sigma_{1}^{2},\sigma_{2}^{2},\sigma_{3}^{2},\sigma_{4}^{2},\sigma_{1}\sigma_{2}^{2},\sigma_{3}\sigma_{2}^{2},\sigma_{1}\sigma_{3}^{2},\sigma_{4}\sigma_{3}^{2}$
are selected. The new structure corresponds to Fig. \ref{fig:new15} and the new weighted linear combination is
\begin{align}
&w_{0,1}+w_{0,2}\sigma_{1}\sigma_{2}+w_{0,3}\sigma_{1}\sigma_{3}+w_{0,4}\sigma_{1}\sigma_{4}+w_{0,5}\sigma_{2}\sigma_{3}+w_{0,6}\sigma_{2}\sigma_{4}+w_{0,7}\sigma_{3}\sigma_{4}+w_{0,8}\sigma_{1}^{2}...\nonumber\\
&w_{0,9}\sigma_{2}^{2}+w_{0,10}\sigma_{3}^{2}+w_{0,11}\sigma_{4}^{2}+w_{0,12}\sigma_{1}\sigma_{2}^{2}+w_{0,13}\sigma_{3}\sigma_{2}^{2}+w_{0,14}\sigma_{1}\sigma_{3}^{2}+w_{0,15}\sigma_{4}\sigma_{3}^{2}
\end{align}
Then, we have
\begin{align}
&\land_1=\varnothing,\land_2=\{1,2\},\land_3=\{1,3\},\land_4=\{1,4\},\land_5=\{2,3\},\land_6=\{2,4\},\land_7=\{3,4\},\land_8=\{1,1\}...\nonumber\\
&\land_9=\{2,2\},\land_10=\{3,3\},\land_{11}=\{4,4\},\land_{12}=\{1,2,2\},\land_{13}=\{2,2,3\},\land_{14}=\{1,3,3\},\land_{15}=\{3,3,4\}
\end{align}
The results of comparative analysis experiments using old and new structure for four-order are also presented, paying particular attention to average error, average best error and average wort correct classification shown in Tables 6, 7 and 8. These lead to verify the theoretical evaluation of SPSNNs learning with new structure is better and monotonically decreasing and converge to 0 during the iterative learning as predicted by Theorem 1.

\begin{figure}[ht]
  \centering
  \includegraphics[width=4in]{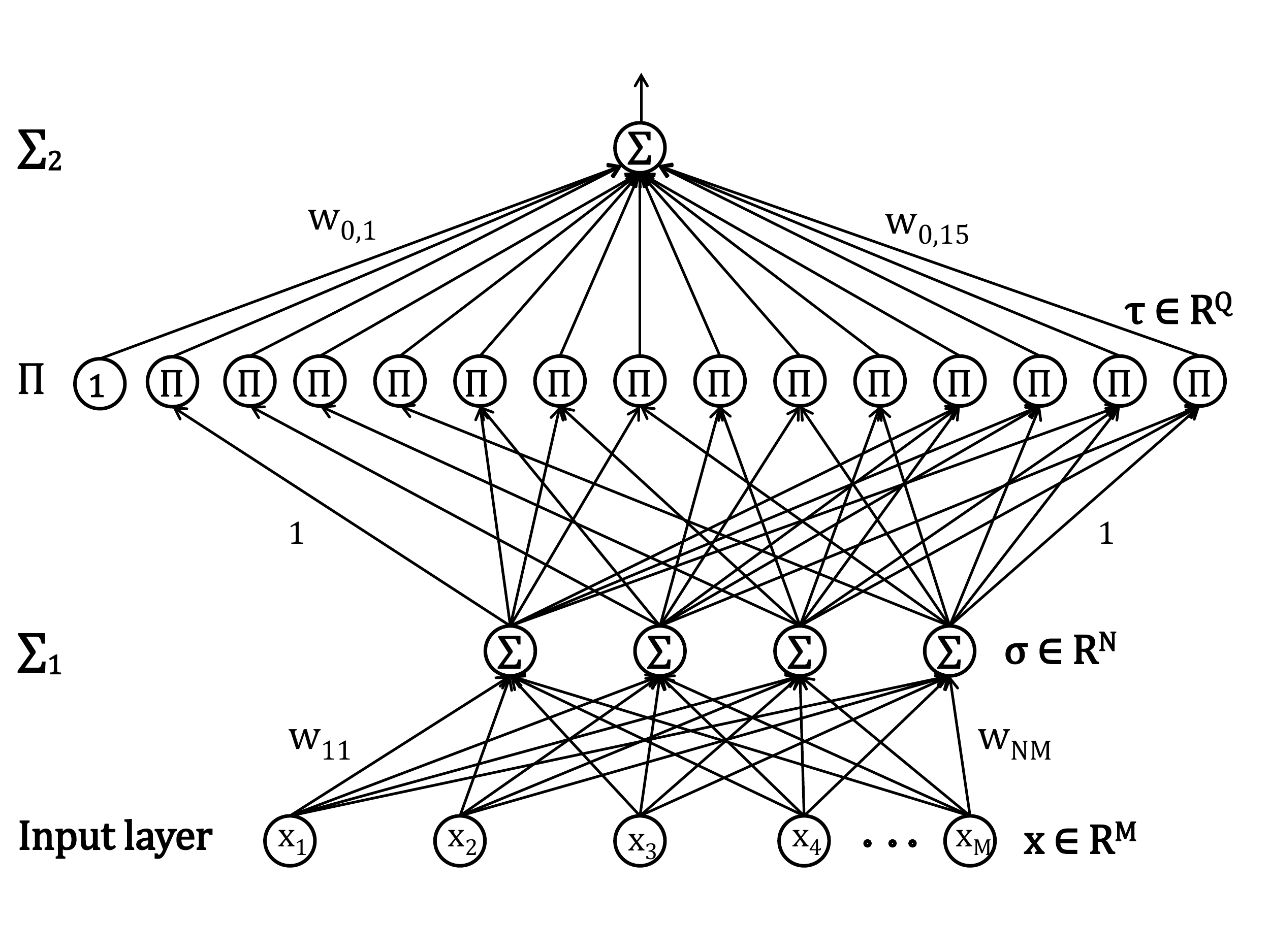}
  \caption{Four-input three-order Old structure }
  \label{fig:old15}
\end{figure}

\begin{table}[!ht]
  \centering
  \caption{Comparison of average error for Pima Indians problem.}
  \label{tab:table6}
  \begin{tabular}{ccccccc}
    \toprule
   Round& Old Train& New Train & Impprovment\% &Old Test & New Test & Improvement\%\\
    \midrule
    1  & 78.76 & 82.53 & 4.67  & 72.31  & 77.57 & 7.02 \\
    2  & 73.04 & 79.38 & 8.32  & 68.24  & 84.62 & 9.2 \\
    3  & 84.33  & 84.70 & 0.44  & 79.44  & 65.38 & 0.24 \\
    4  & 81.01 & 85.81 & 5.75  & 75.86  & 71.79 & 4.18 \\
    5  & 75.97 & 81.22 & 6.68  & 71.52  & 78.21 & 5.89 \\
    \bottomrule
    Overall  & 78.62 & 82.73 & 5.09 & 73.47 & 77.4 & 7.79 \\
    \bottomrule
  \end{tabular}
\end{table}
\begin{table}[!ht]
  \centering
  \caption{Comparison of best error for Pima Indians problem.}
  \label{tab:table7}
  \begin{tabular}{ccccccc}
    \toprule
   Round& Old Train& New Train & Impprovment\% &Old Test & New Test & Improvement\%\\
    \midrule
    1  & 88.13  & 89.39 & 1.42  & 76.67  & 786.61 & 12.18 \\
    2  & 77.01  & 85.97 & 11.0  & 74.58 & 78.75 & 5.44\\
    3  & 89.17  & 91.37 & 2.44  & 85.97  & 88.06 & 2.40 \\
    4  & 87.08  & 90.00  & 3.30 & 81.53  & 83.93 & 2.90 \\
    5  & 78.75  & 83.76 & 6.17  & 75.63  & 79.70 & 5.24 \\
    \bottomrule
    Overall  & 84.03 & 88.10& 4.73 & 78.88 & 83.41 & 5.70 \\
    \bottomrule
  \end{tabular}
\end{table}
\begin{table}[!ht]
  \centering
  \caption{Comparison of  wort error for Pima indians problem.}
  \label{tab:table8}
  \begin{tabular}{ccccccc}
    \toprule
   Round& Old Train& New Train & Impprovment\% &Old Test & New Test & Improvement\%\\
    \midrule
    1  & 73.61 & 75.41 & 2.42  & 69.74  & 71.59 & 2.62 \\
    2  & 63.49 & 73.0 & 13.94  & 61.98  & 66.76& 7.43 \\
    3  & 76.04  & 77.08 & 1.36  & 69.44  & 71.40& 2.78 \\
    4  & 69.44 & 76.04 & 9.07  & 61.98  & 69.44 & 11.35 \\
    5  & 73.96 & 79.44 & 7.2  & 69.44  & 71.98 & 3.59 \\
    \bottomrule
    Overall  & 71.31 & 76.19& 13.0 & 66.52& 70.23 & 5.43 \\
    \bottomrule
  \end{tabular}
\end{table}

\section{Conclusion}
In this study, we use the smoothing $L_{1/2}$ regularization to automatically select some appropriate terms to approximate the complete Kolmogorov-Gabor Multinomial for the product layer of SPSNNs.Numerical experiments are implemented for Mayas' function problem, Gabor function problem, Sonar data classification and Pima Indians diabetes data classification. The numerical results demonstrate that the terms, or equivalently the product nodes selected by using smoothing $L_{1/2}$ regularization have been proved the capability of providing more possibility powerful mapping, which is different from that in old structure.

\end{document}